%% file: main.tex
\documentclass[10pt,conference]{IEEEtran}

\usepackage[group-separator={,}, group-minimum-digits={4}]{siunitx}
\usepackage{pbox}
\usepackage{amsmath}
\usepackage{tablefootnote}
\usepackage{gensymb}
\usepackage{epsfig}
\usepackage{amssymb,amsfonts,latexsym}
\usepackage{enumerate}
\usepackage{xspace}
\usepackage{epsf,picinpar}
\usepackage{varioref}
\usepackage{colortbl,multirow,hhline}
\usepackage{listings}
\usepackage{amssymb}
\usepackage{colortbl,multirow,hhline}
\usepackage{algorithmic}
\usepackage{algorithm}
\usepackage{caption}
\usepackage[normalem]{ulem}
\usepackage{xcolor}
\usepackage{pifont}
\usepackage{xcolor,colortbl}
\usepackage{url}
\makeatletter
\g@addto@macro{\UrlBreaks}{\UrlOrds}
\makeatother
\usepackage{balance}
\usepackage{graphicx, subfigure}
\usepackage{longtable}
\usepackage{lscape}
\usepackage{multirow}
\usepackage{listings}
\usepackage{framed}
\usepackage{morefloats}
\usepackage[T1]{fontenc}
\usepackage{array}
\usepackage{pdfpages}
\usepackage{fancybox}
\usepackage{flushend}
\usepackage{booktabs}
\usepackage{enumitem}
\usepackage{tabularx}
\usepackage{diagbox}
\usepackage{soul} 
\usepackage{pgfplots} 
\usepackage{eurosym}

\usepackage{authblk}

\newcolumntype{P}[1]{>{\raggedright\arraybackslash}p{#1}}

\newcolumntype{H}{>{\setbox0=\hbox\bgroup}c<{\egroup}@{}}
\newcolumntype{Z}{>{\setbox0=\hbox\bgroup}c<{\egroup}@{\hspace*{-\tabcolsep}}}

\definecolor{gray50}{gray}{.5}
\definecolor{gray40}{gray}{.6}
\definecolor{gray30}{gray}{.7}
\definecolor{gray20}{gray}{.8}
\definecolor{gray10}{gray}{.9}
\definecolor{gray05}{gray}{.95}

\definecolor{celadon}{rgb}{0.67, 0.88, 0.69}
\definecolor{lightcoral}{rgb}{0.94, 0.5, 0.5}

\definecolor{black}{rgb}{0.2, 0.2, 0.2}
\definecolor{darkgreen}{rgb}{0, 0.5, 0}

\definecolor{lightgrey}{rgb}{0.9, 0.9, 0.9}

\newcommand{\todo}[1]{\textcolor{blue}{\ding{46}~#1}} 
\newcommand{\ie}{\emph{i.e.,}\xspace}
\newcommand{\eg}{\emph{e.g.,}\xspace}
\newcommand{\etc}{etc.\xspace}
\newcommand{\etal}{\emph{et~al.}\xspace} 

\newcommand{\PreserveBackslash}[1]{\let\temp=\\#1\let\\=\temp}
\newcolumntype{C}[1]{>{\PreserveBackslash\centering}p{#1}}
\newcolumntype{R}[1]{>{\PreserveBackslash\raggedleft}p{#1}}
\newcolumntype{L}[1]{>{\PreserveBackslash\raggedright}p{#1}}

\makeatletter
\newcommand\footnoteref[1]{\protected@xdef\@thefnmark{\ref{#1}}\@footnotemark}

\makeatother

\usepackage{array}
\newcolumntype{P}[1]{>{\centering\arraybackslash}p{#1}}

\newcommand{\patricia}[1]{\patricia{blue}{{\it [Patricia: #1]}}}
\newcommand{\michel}[1]{\michel{red}{{\it [Grace: #1]}}}

\newcommand{\theme}[2]{\vspace{0.2cm}
\noindent \textbf{#1} (#2) --\xspace}

\newcommand\q[1]{\textit{``#1''}}
\newcommand\datapoint[1]{\texttt{#1}}

\definecolor{dkgreen}{rgb}{0,0.6,0}
\definecolor{gray}{rgb}{0.5,0.5,0.5}
\definecolor{mauve}{rgb}{0.58,0,0.82}

\definecolor{mygray}{gray}{1}

\begin{document}

\title{Mining Energy-Related Practices \\ in Robotics Software}



\author[$\dagger$]{Michel Albonico}
\author[$\ddag$]{Ivano Malavolta}
\author[$\mp$]{Gustavo Pinto}
\author[$\ddag$]{Emitza Guzman}
\author[$\ddag$]{\\Katerina Chinnappan}
\author[$\ddag$]{Patricia Lago}
\affil[$\dagger$]{Technological Federal University of Paran\'a - UTFPR, Brazil - michelalbonico@utfpr.edu.br}
\affil[$\ddag$]{Vrije Universiteit Amsterdam, The Netherlands - \{i.malavolta,e.guzmanortega,k.p.chinnappan,p.lago\}@vu.nl}
\affil[$\mp$]{Federal University of Par\'a, Brazil - gpinto@ufpa.br}


\maketitle

\begin{abstract}

Robots are becoming more and more commonplace in many industry settings. This successful adoption can be partly attributed to (1) their increasingly affordable cost and (2) the possibility of developing intelligent, software-driven robots. Unfortunately, robotics software consumes significant amounts of energy. Moreover, robots are often battery-driven, meaning that even a small energy improvement can help reduce its energy footprint and increase its autonomy and user experience. 

In this paper, we study the Robot Operating System (ROS) ecosystem, the \emph{de-facto} standard for developing and prototyping robotics software. We analyze \num{527} energy-related data points (including commits, pull-requests and issues on ROS-related repositories, ROS-related questions on StackOverflow, ROS Discourse, ROS Answers and the official ROS Wiki). 

Our results include a quantification of the interest of roboticists on software energy efficiency, \num{10} recurrent causes and \num{14} solutions of energy-related issues, and their implied trade-offs with respect to other quality attributes.
Those contributions support roboticists and researchers towards having energy-efficient software in future robotics projects.

\end{abstract}


\input{sections/intro}

\input{sections/background}
\input{sections/design}
\input{sections/results}
\input{sections/discussion}
\input{sections/threats}

\input{sections/related} 
\input{sections/conclusions} 

\section*{Acknowledgments}
This research is partially supported by the Dutch Research Council (NWO) through the OCENW.XS2.038 grant; the CNPQ/FA through the PPP-CP-20/2018 call; and the FAPESPA.

\balance


\bibliographystyle{abbrv}
\bibliography{references}

\end{document}

%% file: sections/intro.tex
\section{Introduction}\label{sec:intro}


The intensive use of robots is becoming a successful, commonplace story in several industrial sectors like manufacturing, logistics, delivery, transportation, healthcare~\cite{robotics_roadmap}.
With large players such as ABB, Siemens, and Mitsubishi, it is estimated that the revenue generated from the industrial robotics market worldwide will be 1\num{8.25} billion USD in $2025$~\cite{robots_revenue}.
%
One of the most important reasons for the success of robotics is the possibility of equipping robots with intelligence~\cite{NL_robots}. With intelligent robots, \textit{software} comes into the picture, becoming \textit{the} core aspect in robotics development~\cite{ICSE_SEIP_2020,ICSE_2017_TB}. 

One of the key technological enablers for robotics software development is the 
Robot Operating System (ROS), a communication framework for robotics software modules~\cite{ros}. 
ROS is the \emph{de-facto} standard for developing and prototyping robotics software. 
It supports 162 different types of robots and has a vibrant open-source ecosystem, including \num{6096} publicly-available software packages, \num{9148} ROS Wiki users, and \num{36901} ROS Answers users \cite{ROS_metrics,jss_2019}. 

Robotics software can consume substantial amounts of \textit{energy}. 
For example, the automotive industry alone in the U.S. spends \num{2.4} billion USD on electric energy annually~\cite{chang2012energy} and industrial robots in the automotive industry consume on average \num{8}\% of the total electrical energy of assembly lines \cite{meike2013energy}.
In this context, even a slight energy improvement can lead to great benefits in terms of environmental impact, mission completion time (\eg fewer pauses for recharging batteries), and safety (\eg a flying drone crashing due to poorly-managed energy consumption).
%
Even though there is a relatively rich body of literature on energy efficiency for robotics~\cite{sustainable_se_2020}, it is still unclear how \textit{practitioners} are dealing with energy-related aspects in projects developed in real contexts.

The \textbf{goal} of this paper is to characterize the recurrent \textit{practices} of roboticists in the field, specifically: 
(i) to quantify the interest of practitioners on energy aspects of robotics software and
(ii) to obtain a deeper understanding of the main causes, solutions, and trade-offs of energy-related issues in robotics software. 

In this work, we apply software repository mining techniques targeting (i) developers' discussions on StackOverflow and official technical forums used by ROS developers (e.g., ROS Answers) and (ii) the source code, documentation, commit messages, issues, and pull requests in \num{335} Git repositories containing real open-source ROS systems \cite{ICSE_SEIP_2020}. Out of \num{339563} potentially-relevant data points\footnote{We use the term \textit{data point} as the superclass of any type of mined item, such as a commit message, a GitHub issue, or a discussion on StackOverflow.}, we systematically curate a final set of \num{527} data points where developers discuss, mention, or consider energy in the context of their ROS systems.
We then employ quantitative and qualitative data analysis techniques 
to extract and synthesize roboticists' practices with respect to energy; specifically: how much roboticists consider energy-related issues in their projects, \num{10} recurrent \textit{causes} and \num{14} \textit{solutions} of energy-related issues, and their implied \textit{trade-offs} with respect to other quality attributes like reliability and performance. 
%
%
The \textbf{main contributions} of this study are:
\begin{itemize}
    \item a quantification of the interest of practitioners in the energy aspects of robotics software;
    \item a taxonomy of the main causes and solutions of energy-related issues faced by roboticists;
    \item the identification of the quality attributes considered by roboticists when facing energy-related issues;
    \item a manually curated and validated dataset of \num{527} energy-related data points in the context of robotics software;
    \item the complete replication package of the study.
\end{itemize}

The \textbf{target audience} of this study includes both roboticists and researchers in green software engineering.  
Roboticists can benefit from the taxonomy of energy-related issues by using the extracted causes as a checklist of the various aspects of their system to be taken under consideration and by using the extracted solutions as a catalog of concrete solutions in case their system suffers from an energy-related issue.
Researchers get an objective characterization of the state of the practice about energy-related aspects in robotics software. 
Moreover, researchers can use the extracted causes and solutions as a foundation for defining new approaches to automatically improve the energy efficiency of robotics software.

%% file: sections/background.tex
\section{Background}\label{sec:background}
\noindent \textbf{The ROS Community} -
The ROS community is rapidly growing and especially active; when a developer encounters a problem, finding solutions and getting help becomes easier -- not only from developers of the ROS platform but also from other enthusiasts and professionals. 
The primary communication channels and resources used in the ROS community are:

\begin{itemize}
    \item[\textbf{i)}] Stack Overflow and ROS Answers: Q\&A websites which are used as one of the main communication channels for solving problems;
    \item[\textbf{ii)}] ROS Discourse: a discussion forum mainly used for discussing more complex topics and announcing new projects and updates concerning the ROS community;
    \item[\textbf{iii)}] GitHub: a repository hosting service where the majority of ROS packages are hosted at and which provides a \textit{Pull Request} (PR) and \textit{Issue} section of the package repository where questions and bugs concerning the ROS package can be discussed, and;
    \item[\textbf{iv)}] ROS Wiki: a collection of various ROS tutorials, packages, and libraries. 
\end{itemize}

Every day members part of the ROS community is heavily involved in the open-source development of publicly available ROS packages. As of July 2020, there are a total of \num{16044} packages published on the official ROS website~\cite{jss_2019}. According to ROS Community Metrics Report for July 2020~\cite{ROS_metrics}, there are roughly \num{142000}, \num{199000}, and \num{21400} registered users respectively for ROS Answers, ROS Wiki, and ROS Discourse.

\vspace{0.2cm}
\noindent \textbf{ROS-based Systems} -
A ROS-based system is composed of \textbf{Nodes} which are OS processes that perform computations~\cite{quigley2015programming}. A registered Node can interact with other Nodes using a publish/subscribe model based on \textbf{Topics} (publish and subscribe to messages), or using a request/response model based on \textbf{Services} or \textbf{Actions} (request and receive responses). A Service is a communication model that operates on the principle of synchronous bidirectional communication between a Service Client that requests data and a Service Server that responds to requests. Service calls are blocking so they are typically used for remote procedure calls that terminate quickly (e.g., simple calculations, querying the state of a Node)~\cite{hernandez2015}. The Action communication model is used when the requested task takes a long time to complete and feedback from the process is needed (\eg moving the robot). 



%% file: sections/design.tex
\section{Study Design}\label{sec:design}

\noindent We designed this study by following known empirical guidelines~\cite{shull2007guide}\cite{wohlin2012experimentation}.
A full replication package is publicly available for independent verification and replication \cite{replication_package}.

\subsection{Goal and Research Questions}\label{sec:rq}
The \textbf{goal} of this study is to 
analyze \textit{the ROS software ecosystem} 
for the purpose of \textit{quantifying and characterizing} 
the main \textit{causes, solutions, and eventual trade-offs of roboticists' issues} 
with respect to their \textit{energy efficiency} 
from the point of view of \textit{roboticists and researchers} 
in the context of \textit{open-source ROS-based systems}. The goal drives the design of the full study and leads us to the following research questions.

\noindent \textbf{RQ1} -- \textit{To what \textbf{extent} do roboticists consider \textbf{energy consumption} in the context of robotics software?}
By answering this research question we aim at quantifying the interest in energy consumption of roboticists (i) over time and (ii) across different types of robotic systems and capabilities.

\noindent \textbf{RQ2} -- \textit{What are the main \textbf{causes} of energy-related issues reported by roboticists?}
This research question targets energy-related issues. In this study,  \textit{\textbf{energy-related issues}} are defined as errors, bugs, faults, or failures affecting the energy consumption of a robotic system, either purposely (\eg streaming a high-definition video to the Cloud) or accidentally (\eg a software bug leading to unnecessary CPU cycles). 

\noindent \textbf{RQ3} -- \textit{What are the main \textbf{solutions} that roboticists apply or recommend for solving energy-related issues?}
This research question is the counterpart of RQ2. Specifically, RQ3 aims at identifying and characterizing the most prevalent solutions either applied or reported by roboticists for solving energy-related issues.


\noindent \textbf{RQ4} -- \textit{What are the \textbf{quality attributes} mentioned by roboticists when considering energy-related issues?}
As ROS-based systems are becoming more and more large and complex, it is critical that their software meets quality requirements such as maintainability, safety, and reliability \cite{ICSE_SEIP_2020}.

\subsection{Dataset Building}\label{sec:dataset}

\begin{figure}[!h]
\includegraphics[width=0.47\textwidth]{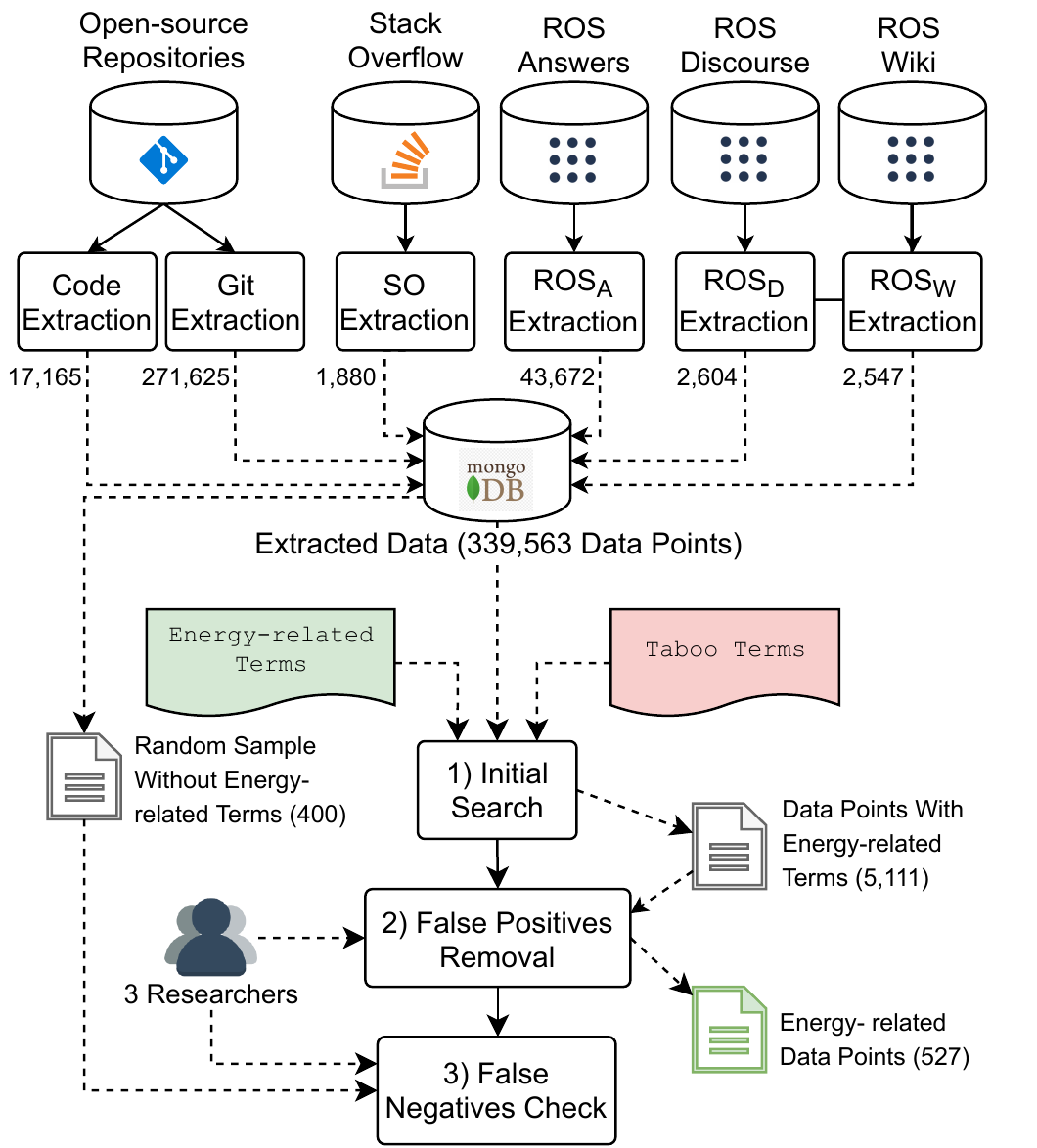}
\centering
\caption{Overview of the dataset building process.}
\label{img:dataset}
\end{figure}

Figure~\ref{img:dataset} gives an overview of the process we followed for building the dataset for answering our RQs. 
We consider the following starting data sources: \emph{Open-source Git repositories}, \emph{Stack Overflow}, \emph{ROS Answers}, \emph{ROS Discourse}, and \emph{ROS Wiki}.
For the open-source repositories, we consider the GitHub, Bitbucket, and Gitlab social coding platforms.
In this context, we reuse an already-existing dataset of \num{335} repositories containing real and active ROS-based projects \cite{ICSE_SEIP_2020}.
We locally clone each repository and extract all its source code comments and markdown files (\emph{Code Extraction} in Figure \ref{img:dataset}).
We also mine all issues/pull requests (including their discussions) and commit messages (\emph{Git Extraction} in the figure).
For \emph{Stack Overflow}, \emph{ROS Answers}, and \emph{ROS Discourse} we crawl all questions, their answers, comments, and related (meta-)data.
Since \emph{Stack Overflow} is not specific to ROS, we target questions with the \texttt{ROS} tag. 
For \emph{ROS Wiki}, we crawl all its pages and related (meta-)data.
All extracted data points are persisted in a MongoDB database. Since MongoDB is schemaless (\ie there are no restrictions on the structure of the stored data), it simplifies the persistence and querying of our (highly-heterogeneous) data points. For the sake of space, the data extractors are detailed in our replication package, together with their source code and a complete dump of the MongoDB database \cite{replication_package}. 
The dataset building activities were carried out in February-April 2020 and result in \num{339563} distinct data points.

\vspace{-4mm}
\vspace{-3mm}

Figure~\ref{img:dataset} also illustrates the three phases we follow to select energy-related data points from our dataset: 1) initial search; 2) false positives removal; and 3) false negatives check. 

\noindent
\textbf{Phase 1 -- }In this phase, we select data points that contain energy-related terms, such as battery, power, energy, sustainability, \etc 
Despite its simplicity and low computational requirements, the keyword-based strategy is commonly used \cite{bavota2016mining} and has been successfully applied in previous studies on mining software repositories and Q\&A platforms about energy efficiency \cite{chowdhury2016characterizing}\cite{cruz2019catalog}\cite{malik2015going}\cite{moura2015mining}.
In this phase, it is fundamental to identify the search terms that best fit with the problem at hand \cite{bavota2016mining}. Two researchers identify the search terms via a semi-systematic process where they: 
(i) collect a set of 14 scientific publications containing a keyword-based search applied in the context of green software across several domains and different types of targeted data (see  Table~\ref{tab:paperswithterms}), (ii) extract all the search terms used in each publication, (iii) merge and adapt the extracted search terms according to the robotics domain (\eg we merge the ``\texttt{energy efficiency}'' term used in \cite{moghaddam2018self} since it is dominated by the ``\texttt{energy}'' term used in \cite{pinto2014mining}). A third researcher performs a final sanity check of the identified search terms. 
The final set of used search terms is:

\vspace{1mm}
\noindent
\fbox{\parbox{0.48\textwidth}{
\emph{*battery*, *energy*, *power*, *sustainab*, *green*, *consum*, *efficien*, *drain*, *sleep*, *charg*, *volt*, *wak*, *watt*, and *joule*}
}}

Finally, we query our MongoDB database by considering the list of energy-related search terms.
After an initial inspection of the obtained search hits, we noticed that we were having several false positives. For instance, the term \texttt{green} leads to matches including forms like \texttt{green LED}, \texttt{green button}, that are clearly out of scope for this study.   
To mitigate this threat, two researchers randomly sampled a subset of the search results and collaboratively identified a set of combinations of terms that are out of scope, we call them \textit{taboo terms}. Examples of taboo terms are \texttt{coefficient}, \texttt{time consuming}, and \texttt{green icon}. Then, we filter out all the search results matching with at least one taboo term.     
%
This results in \num{5111} data points containing energy-related terms.
The high discard-rate in this step is not surprising,
but is rather in accordance with existing research confirming that developers tend to have limited knowledge of energy efficiency \cite{pang2015programmers}.

\vspace{-1mm}
\begin{table}[h]
\centering
\caption{Studies used for extracting energy-related terms.}
\begin{tabular}{ l p{1.1cm} p{2.5cm} }
 \hline
 \textbf{Paper} & \textbf{Domain} & \textbf{Targeted Data} \\ 
 \hline

  Li et al.~\cite{li2020detecting}& Mobile & Issues \\
  Swanborn and Malavolta~\cite{sustainable_se_2020} & Robotics & Literature \\
 Cruz and Abreu~\cite{cruz2019catalog} & Mobile & Commits, Issues, PRs \\
 Moghaddam, Lago, and Ban~\cite{moghaddam2018self} & Generic & Literature \\
 Matalonga et al.~\cite{matalonga2019greenhub} & Mobile & System Events \\
 Chowdhury and Hindle~\cite{chowdhury2016characterizing} & Generic & Commit Messages \\
 Bavota~\cite{bavota2016mining} & Generic & Commit Messages \\
 Moura et al.~\cite{moura2015mining} & Generic & Commits \\
 Malik, Zhao and Godfrey~\cite{malik2015going} & Generic & Stack Overflow  \\
 Procaccianti, Lago and Bevini~\cite{procaccianti2015systematic} & Cloud & Literature \\
 Pinto, Castor and Liu~\cite{pinto2014mining} & Generic & Stack Overflow \\
 Procaccianti, Bevini and Lago~\cite{procaccianti2013energy} & Cloud & Literature \\
 Calero, Bertoa and Moraga~\cite{calero2013systematic} & Generic &  Literature \\
 Pathak, Hu and Zhang~\cite{pathak2011bootstrapping} & Mobile & Issues, Forum Posts\\
 \hline
\end{tabular}
\label{tab:paperswithterms}
\vspace{-2mm}
\end{table}

\noindent
\textbf{Phase 2 -- } 
In this phase we \textit{manually} analyze all the \num{5111} data points for removing false positives.
This phase is performed systematically and iteratively by three researchers.
In a first iteration, we perform a stratified random sampling of \num{398} data points according to their type; specifically, we randomly sample \num{50} data points for each type of data point (\ie 50 Git commits, 50 Stack Overflow discussions, and so on). Then, two researchers independently assess whether each sampled data point is a false positive. We verify the inter-rater agreement via the Cohen's Kappa coefficient~\cite{cohen1968weighted}, which is \num{0.82}, thus resulting in an almost perfect agreement.  %
Then, we discuss and solve the occurred conflicts with the help of a third researcher acting as arbiter.
Subsequently, we repeat the same process by considering another random sample of \num{50} data points for each type of data data point; this time the Cohen's Kappa coefficient increases to \num{0.84}, making us confident about the objectivity of our manual classification. Based on the good inter-rater agreement obtained in the first two iterations, one researcher proceeds to classify the remaining data points, with the help of another researcher in case of doubts.
This phase results in \num{527} energy-related data points. 


\noindent
\textbf{Phase 3 -- }
When performing a keyword-based search, it is fundamental to minimize the number of false negatives \cite{bavota2016mining}, \ie data points that do \textit{not} contain any of energy-related terms in our list but are about energy. 
In this context, we are interested in the data which have not been selected in Phase 1 (\begin{math}\num{339563} - \num{5111} = \num{334452}\end{math}). Since a manual analysis of all \num{334452} is not feasible, we consider a random sample of \num{400} data points, stratified again according to the \num{8} types of data points (\num{50} data points for each type). By considering \num{400} data points, we achieve a \num{95}\% confidence interval with a \num{4.9}\% error margin (assuming a 50\% population proportion). 
We classified the \num{400} by following the same process as in Phase 2. 
This process resulted in the identification of zero false negatives, making us reasonably confident about having an acceptable number of false negatives in our study.



\subsection{Data Analysis}\label{sec:data_analysis}

The selected energy-related data points are analyzed in three different ways in order to answer the research questions, where we follow the same strategy to answer \emph{RQ2} and \emph{RQ3}.
All the data analysis is detailed in the remainder of this section.

\noindent \textbf{RQ1} -- Firstly, we collect the creation date of each of the 527 data points in our dataset. Then, two researchers conduct iterative content analysis sessions with open coding \cite{lidwell2010universal} to extract information related to (i) the considered types of robots (\eg ground rover, flying drones, industrial arms) and (ii) their provided capabilities (\eg vision, navigation, manipulation). Then, we apply simple summary statistics to answer RQ1 quantitatively. 

\noindent \textbf{RQ2 and RQ3} -- 
These research questions are answered by applying \textit{thematic analysis} \cite{fereday2006demonstrating,thematic_analysis,vaismoradi2013content}.
Thematic analysis is a qualitative research method for identifying emerging patterns from a body of knowledge.
In our study, the body of knowledge is the set of \num{527} energy-related data points, and the emerging patterns are used to answer RQ2 and RQ3.
We chose thematic analysis because (i) it has been successfully applied in several previous studies on energy-efficient software, \eg \cite{moura2015mining,cruz2019catalog} and (ii) the information we extracted from each data point is strongly dependent on project- and system-specific characteristics and thematic analysis copes well with context-dependent data \cite{vaismoradi2013content,thematic_analysis}.
By following the guidelines reported in \cite{fereday2006demonstrating}, three researchers carried out the thematic analysis according to the following steps:   





\noindent
\textit{1) Familiarisation with the data --} The three researchers carefully inspect all \num{527} data points to become familiar with the dataset. 
When a data point is about concepts/techniques which are not common or unclear, we study them using either related scientific literature or online documentation (usually via the official ROS Wiki).

\noindent
\textit{2) Extracting initial codes --} 
The goal of this step is to extract descriptive labels from segments of text from each data point. Examples of extracted labels are: ``\texttt{high modulation frequency}'', ``\texttt{mpeg\_server is power consuming}'', ``\texttt{land when battery is low}''.
We extracted the initial codes in five main steps. In the first one, we randomly sampled \num{80} data points (\num{10} for each type -- see Table \ref{tab:dpointsdist}) and three researchers code them in parallel; in the second step, the extracted codes are largely discussed in order to identify differences of perspectives and spot potential sources of subjectivity. No substantial differences were identified in the extracted codes. In the third step one researcher proceeds with extracting the codes of the remaining data points, with the help of the other two researchers in case of doubts. Once all the codes are extracted, in the fourth step we collaboratively and iteratively combine codes with the same meaning, resulting in a final set of \num{494} unique codes.
Finally, in step 5 we group the extract codes according to the research questions they may answer, leading to \num{158} codes related to \emph{RQ2} and \num{307} codes related to \emph{RQ3} (with 24 codes belonging to both groups). Contextually, we discard \num{52} codes that are not related either to RQ2 or RQ3.



\noindent
\textit{3) Searching for themes -- } The goal of this step is to combine the extracted codes into an initial set of themes. 
For each RQ, two researchers navigate through the extracted codes and organize them into meaningful subsets. 
Then, for each subset, a precise theme was formulated. We focused on making the phrasing of the themes (i) representative of its corresponding codes, (ii) not overlapping, and (iii) actionable to be readily usable by roboticists in the field. This activity resulted in a total of \num{28} themes for RQ2 (the causes of energy-related issues) and \num{41} themes for RQ3 (their solutions).


\noindent
\textit{4) Reviewing themes --} In this step we go back to the extracted codes once again in order to rearrange the codes and refine themes.
Here, generic themes are split into more specific subcategories, others are renamed (when necessary), and codes are moved from one theme to another (when necessary).
While reviewing the codes, we search for data that supports or refutes themes.
In most cases, we only go back to the extracted codes. 
When the extracted codes are not descriptive enough, 
we go back to the initial data points.
This activity resulted in a total of \num{24} themes,  \num{10} themes for RQ2 and \num{14} themes for RQ3. 

\noindent
\textit{5) Defining and naming themes --} 
In this step, two researchers finalize the name and produce a structured description of each of the \num{24} themes. The description of each theme is based on the data points in our dataset (\eg a solution described in a GitHub pull request), the researcher's experience, logical arguments, and the scientific literature. Finally, the phrasing (and semantics) of each theme undergo rigorous scrutiny of one additional researcher, leading to a further refinement of the emerged themes.

\noindent
\textit{6) Producing the final report --} 
In this phase, all authors of this study carefully scrutinize the emerged themes, contextualise them, and further refine their definitions.

\noindent \textbf{RQ4} --
For our last research question, we revisit all \num{527} data points that have been qualitatively analyzed to answer RQ2 and RQ3. In this second round of assessment, we revisit the data points focusing on quality attributes mentioned in the discussion/code change. We use the quality attributes defined in the ISO/IEC 25010 standard \cite{ISO25010} as the initial set of codes, which encompasses eight main groups, namely: Functional Stability, Performance, Compatibility, Usability, Reliability, Security, Maintainability, and Portability. For each data point, we verify if there is any mention of these quality attributes. Ultimately, we conduct content analysis sessions on the data points mentioning at least one quality attribute and report the frequency in which each quality attribute is mentioned contextually to energy, together with a description of notable/representative examples. 

%% file: sections/results.tex

\section{Results}\label{sec:results}


\subsection{Consideration of energy-related issues (\textbf{RQ1})}\label{sec:results_rq1}
Energy-related discussions are rare in the analyzed artifacts. We discovered \num{527} (0.002\%) energy-related data points out of our initial dataset of 339,563. The most common types of energy-related data points were ROS Answers discussions (\num{30.93}\%), Git commit messages (\num{26.19}\%), Git issues (\num{22.01}\%), and Git pull requests (\num{11.76}\%). 

\vspace{-1mm}
\begin{table}[!h]
\centering
\vspace{-1mm}
\caption{Number of energy-related data points.}
\begin{tabular}{ l r r r  }
\hline
 \textbf{Type of data point} & \textbf{All} & \textbf{Energy-related} & \textbf{Pct.} \\ 
 \hline
 \textbf{ROS Answers discussions} & \num{43672} & \num{163} & \num{30.93}\%\\
 \textbf{Git commit messages} & \num{218385} & \num{138} & \num{26.19}\% \\
 \textbf{Git issues} & \num{23214} & \num{116} & \num{22.01}\%\\
 \textbf{Git pull requests} & \num{30096} & \num{62} & \num{11.76}\% \\
 \textbf{Source code} & \num{16069} & \num{18} & \num{3.42}\% \\
 \textbf{ROS Discourse discussions} & \num{2604} & \num{17} & \num{3.22}\% \\
 \textbf{ROS Wiki pages} & \num{2547} & \num{12} & \num{2.28}\% \\
  \textbf{Stack Overflow discussions} & \num{1880} & \num{1} & \num{0.19}\% \\
  \hline
                \textbf{TOTAL}         & \num{339563} & \textbf{527} & 100\% \\
\end{tabular}
\label{tab:dpointsdist}
\end{table}
\vspace{-1mm}

\begin{figure}[!h]
\includegraphics[width=\columnwidth]{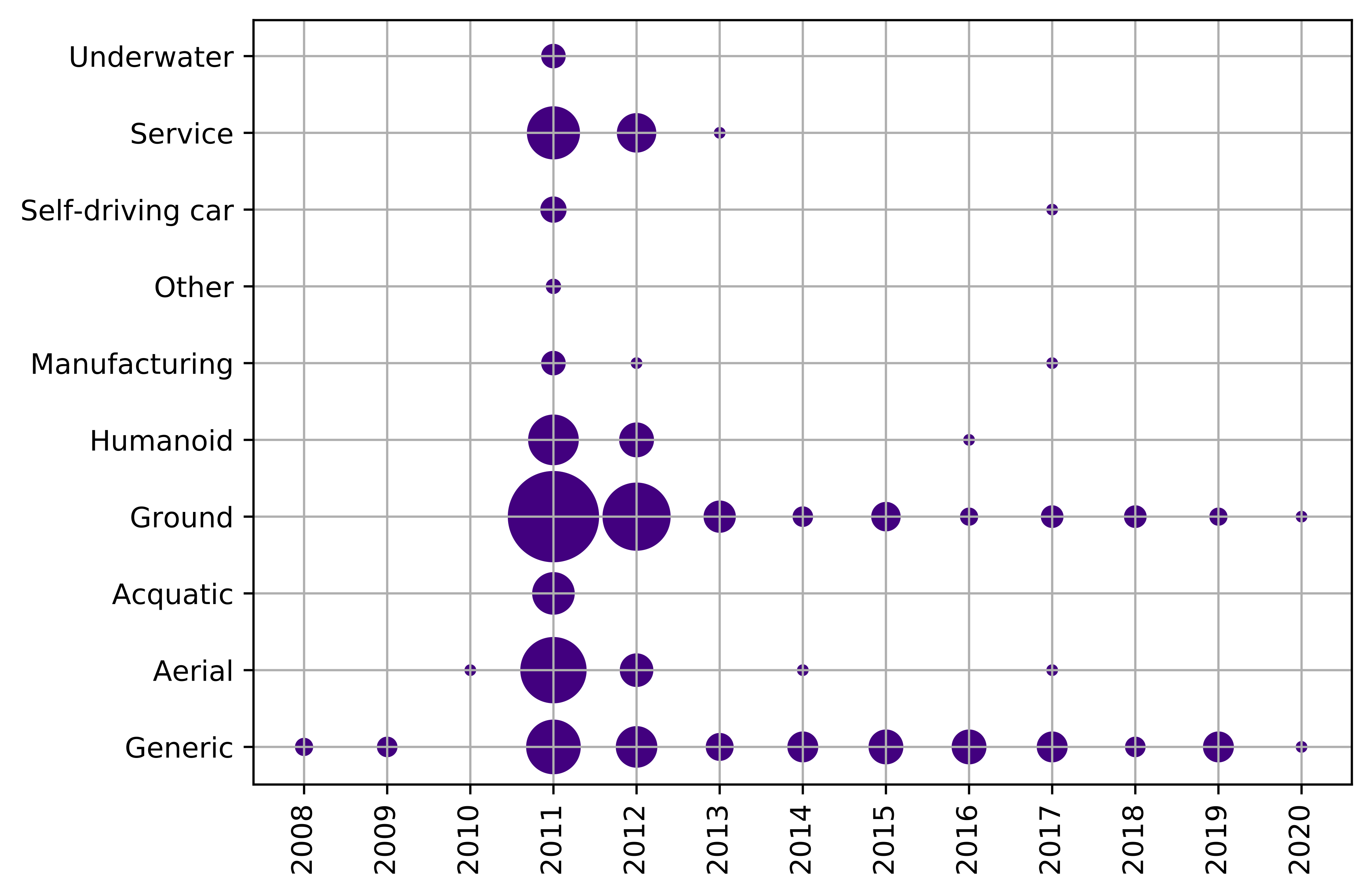}
\centering
\vspace{-5mm}
\caption{Energy-related data points over time per robot type.}
\label{img:overtime}
\end{figure}

Table~\ref{tab:dpointsdist} shows the distribution of energy-related data points among all considered types of data points. Most of these data points are associated to ground (36.8\%), generic (24.5\%), and aerial (12.5\%) robots. Most of these energy-related data points are associated to robots with full (35.5\%), base\footnote{Subsystem containing the base components of the whole ROS system} (25\%), and infrastructural (10.1\%) capabilities. The most common energy-related terms in our dataset were \textit{*battery*} (45.73\%), \textit{*power*} (15.75\%) and \textit{*charg*} (12.58\%). Figure~\ref{img:overtime} shows the energy-related data points overtime per robot type in the time span between 2008-2020. The years 2011 and 2012 showed considerable more energy-related discussions than the rest of the considered years. The figure also shows that ground robots were the most prevalent in these two popular years, albeit their popularity decreased significantly in the upcoming years.

\vspace{-4mm}
\vspace{-3mm}

\subsection{Main causes of energy-related issues (\textbf{RQ2})}\label{sec:results_rq2}
Out of the \num{527} energy-related data points, we identified a total of \num{109} (\num{20.68}\%) distinct ones referring to \textit{causes} of energy-related issues. Then, we further analyzed those \num{109} data points and extracted \num{10} recurrent themes. For the sake of space, in the following, we describe the \num{5} most frequent themes, together with concrete examples and quotations helpful for understanding the context in which energy-related issues may arise in ROS-based systems.
 
\theme{Battery physical properties}{\num{29}}
Robots can be equipped with batteries with widely different operating principles, voltage, and form factors, ranging from standard laptop batteries to sodium-ion batteries, micro-scale lithium-ion batteries, \etc 
As emerged in our analysis, not taking into account those properties can lead to energy-related issues such as unexpected drops in the voltage provided by the batteries, which can lead to a sudden halt of the system, or fast energy draining. For example, in \datapoint{DP221} we can get a clear understanding about the types of battery-related issues a roboticist might have to deal with: \q{Kinect can actually handle a small voltage swing pretty well. I had no problems with \num{11} to \num{13.5}V although I would advise against it. Using a typical 7812 or LM1084-12 is the best option. Also, an ATX output provides a stable \num{12}V (sometimes they do need a minimal load). The Kinect PSU is rated for \num{1} or \num{1.5}A, but using the Kinect with camera + pointcloud only, I never got it above \num{500}mA. However, the current peaks can bring the robot down}. 

Moreover, the physical properties of batteries degrade over time, even when they are not used.
For example, in \datapoint{DP410} a roboticist mentions that \q{if you charge the [Kobuki] batteries and then unplug them, they should survive 6 months to a year. Open the bottom of the Kobuki and unplug the battery pack. And remove the battery from the laptop, like it was shipped.}. These specific types of energy-related causes are especially cumbersome since they might lead to transient bugs, which are difficult to reproduce while testing the system. 

\theme{Bugs and technical issues in the source code}{\num{24}}
The majority of data points in this family are about \textit{bugs} in the source code, which leads to energy-related issues. For example, in \datapoint{DP376}, a roboticist refers to a bug in the Onboard SDK which has been fixed in the \num{3.9} version with the following release note: \q{Battery Information ROS Topic: Fix bug and implement the battery information of ROS topic} \cite{BibEntry2020May}. 

Energy bugs can escalate into failures impacting the whole robotic mission. In \datapoint{DP218} the port used by the controller of a robot was not correctly specified in the ROS parameter server, leading to the situation where \q{the battery [of a Pioneer \num{3}-DX robot] is 0\% even though the robot is fully charged. [The developer is] not able to enable or disable the motors.}. Similar failures can be caused by other types of bugs, such as erroneous management of physical units; \eg in \datapoint{DP347}, battery charge was provided in mAh by the driver, but its corresponding ROS node was publishing it as Ah. The problem of physical units inconsistencies is recognized in the literature and there are approaches for their automatic detection, \eg via static analysis and probabilistic inference~\cite{kate2018phys}. 

Other causes of code-related energy issues include technical debt (\datapoint{DP320}), and the difficulty of developing embedded software for novice developers, which might involve cross-compilation and manual deployment of binary code  (\datapoint{DP6}).

\theme{ROS nodes configuration}{\num{17}}
From an architectural perspective, ROS-based systems are composed of independently-deployed nodes mostly communicating asynchronously via publish/subscribe messages. If on one side the ROS architectural style supports developers in terms of maintenance and portability \cite{ICSE_SEIP_2020}, on the other side in our dataset we observe that configuration errors of ROS nodes can lead to energy-related issues. 
In several data points, the root cause of those configuration errors is located in the launch file of the system, where roboticists either did not correctly remap topics or specified erroneous arguments when declaring ROS nodes. These types of problems can lead to severe consequences. For example, in \datapoint{DP346}, the ROS topic for the battery voltage is not visible at all within the system, potentially leading to ROS nodes subscribing to topics that will never produce any data. 
Moreover, such problems might lead to missing system-wide diagnostics information, thus potentially leading to safety or reliability issues. For example, in \datapoint{DP388} \q{The diagnostics interface was written for/around the voltage reporting abilities of the V2 [of a Segway-based robot]. The V4 segbots have much more rich battery information (State of Charge for both the propulsion and auxiliary battery) which is not tracked by Hardware Diagnostics. Additionally, sensors such as the Velodyne VLP-16, sensors specific to the V4, are not tracked by Diagnostics.}
In other cases, the configuration of the system is done without taking into consideration the execution environment of the system. For example, in \datapoint{DP316}, the ROS node in charge of providing information about the battery charge of the laptop continuously publishes warning messages to other nodes, even if the battery is not present (\ie the laptop is powered directly via its power adapter); this misconfiguration results in a high number of irrelevant messages published during the whole duration of the robotic mission, thus leading to a severe communication overhead.   

\theme{Robot navigation}{\num{11}}
As shown in Section \ref{sec:results_rq1}, mobile robots like rovers and drones are particularly discussed in the ROS ecosystem. 
As confirmed by the literature \cite{sustainable_se_2020}, one of the main energy consumers in mobile robots is robot \textit{navigation} (\eg robot acceleration, deceleration, turns). This phenomenon can be traced back to the fact that mobile robots heavily interact with hardware components that are particularly heavy on the battery, such as servos, motors, \etc. For example, in \datapoint{DP384} the well-known \texttt{ros-planning/navigation2} ROS package is extended with a new feature for configuring the maximum speed of the robot at runtime is proposed as a way to reduce the energy consumption of the robot.  

\theme{Execution environment (\eg board, OS, SDK)}{\num{11}}
The execution environment of the ROS nodes is also influencing the energy consumption of the controlled robot. This phenomenon can happen at different levels of abstraction, ranging from the board where the ROS nodes are running (\eg \datapoint{DP6}), to the OS (\eg \datapoint{DP198}) or the used Software Development Kit -- SDK (\eg \datapoint{DP376}). As an example, the discussion in \datapoint{DP48} includes a suggestion about recommended Single-board Computers (SBC) for ROS development saying that roboticists \q{need an SBC with at least a \num{1.0}Ghz modern CPU + \num{2}GB of RAM. If power consumption is not a concern, [the roboticist] would recommend Intel Core CPUs over Atoms.}. This example highlights the fact that the choice of used processor impacts the energy consumption of the robot.

\theme{Other notable themes}{\num{40}}
Other notable causes of energy-related issues include: streaming images/videos (\num{10} occurrences), interaction with sensors and actuators (\num{9} occurrences), providing feedback to a GUI (\num{8} occurrences), raw computation (\num{7}\ occurrences), and networking (\num{3} occurrences). Overall, the emerged themes are expected and most of them have been already identified as sources of energy-related issues in the literature, both for robotics software \cite{sustainable_se_2020} and other domains like mobile apps \cite{pinto2017energy,pinto2014mining,hao2013estimating}. This result of our study confirms that roboticists are facing in practice similar causes as in other application domains.



\subsection{Main solutions for energy-related issues (\textbf{RQ3})}\label{sec:results_rq3}

From the \num{494} distinct codes emerged in this study, we classify \num{303} as a \emph{solution}.
Those codes cover \num{380} (\num{72.11}\%) data points out of the \num{527} we select as energy-related, which results in \num{14} distinct solution themes.
Due to space limitations, in this study, we describe the \num{6} most recurring solutions (they describe together \num{85.54}\% of the solution data points). 


\theme{Energy monitoring and tracking}{\num{166}}
Most of the solutions that we found are discussions or implementations of monitoring and tracking of energy parameters.
This phenomenon can be related to the fact that robots, likewise other battery-based devices, need to be aware of their battery conditions, such as their charge and remaining operational time. 
Autonomous robots also self-adapt based on battery information, where their behavior is updated in order to save energy when the battery level is low, or even favor processing when the battery level is not a constraint~\cite{MSR_2021_architectural_tactics}.
Therefore, monitoring and tracking energy parameters is a key element when dealing with energy efficiency.

Data point \datapoint{DP111} gives a concrete example of the importance of energy monitoring in a robotics system: \q{the robot monitors its battery level on a topic called /battery\_level using a SMACH MonitorState which is part of the state machine SM\_MONITOR\_BATTERY. [...]
If the battery level falls below a threshold (SM\_MONITOR\_BATTERY returns ''invalid''), SM\_MONITOR\_BATTERY transitions to another state called RECHARGE that moves the robot to the docking station (NAV\_DOCKING\_STATION) and recharges the robot.}
We see that the robot relies on the energy information delivered in a ROS topic by a dedicated battery monitor.
In this case, without the battery information, the robot cannot go back to its docking station to self-recharge, thus potentially affecting the success of the robotic mission.

Voltage and current are the most commonly monitored metrics.
In addition, the internal battery temperature is also considered in \num{4} data points. Battery temperature is supported by the ROS Smart Battery System (SBS); data point \datapoint{DP2} shares the ROS SBS specifications as an attachment that considers the internal battery temperature as an important parameter for a safety check.
For instance, in \datapoint{DP2} a temperature alarm is \q{set when the Smart Battery detects that its internal temperature is
greater than a preset allowable limit. When this bit is set, charging should be stopped as soon as possible.}
This can avoid battery damage and keep its physical characteristics fit, which may increase battery lifetime, and as consequence, make robots more energy-efficient.

\theme{Adaptations in robot behavior}{\num{101}}
Several solutions rely on the robot adapting itself in order to respond to (excessive) energy consumption. 
We find three (non mutually-exclusive) adaptations that are common among our data points: 

\noindent $\bullet$ \emph{Abort mission:} 
the robot stops what it is doing due to energy restrictions.
As an example, data point \datapoint{DP84} is about a flying drone where the energy level of the drone is the main factor for deciding whether the drone should fly or not. In this specific case, if the battery level is below a given threshold, the drone should try to land (both while taking off or when traveling towards a certain location).

\noindent $\bullet$ \emph{Shutdown:}
the robot shuts down itself and the other parts that hold it.
In data point \datapoint{DP103}, a roboticist asks others how to shut down the whole system when the battery level reaches a certain threshold.
Data point \datapoint{DP176} illustrates another case, where another roboticist goes further and asks how to make this decision based on ROS messages.
For \datapoint{DP103}, others suggest creating a service with superuser rights that can be invoked to turn the whole system off. 

\noindent $\bullet$ \emph{Limit robot capabilities:}
once the energy level is low, the robot decreases the resources allocated to some capabilities or even disables those that are not vital for the completion of the mission.
For example, \datapoint{DP50} suggests to trade-off image resolution and frame rate to reduce energy consumption; in normal operations, the robot publishes high-quality images, but it decreases their quality when the battery is below a certain threshold. 
Data point \datapoint{DP159} illustrates a situation where the Kinect\ device is completely disabled in order to save energy.
This can be used either when a sensor has not been used for a while or when the robot can switch to another less energy-consuming sensor, in case precision can be traded off.
Both solutions present themselves as pertinent strategies when the robot depends on or supports human intervention. In those cases, the robot sends a notification to the roboticist for informing them that (i) it is running out of battery and (ii) it is working with limited features.
This gives the roboticist some time to react; and then, replace batteries, or recharge them.
 
\noindent $\bullet$ \emph{Other adaptations:}
Our study also reveals other less frequent adaptations related to energy efficiency:
\begin{itemize}
    \item Go back home (\eg \datapoint{DP423}): the robot returns to is dock station or the starting point of the mission;
    \item Charge during the mission (\eg \datapoint{DP18}): the robot keeps performing the mission, while its batteries are recharged;
    \item Hibernation (\eg \datapoint{DP5}): the robot enters a hibernation state until its batteries are recharged.
\end{itemize}

\theme{User interfaces for assisting operators}{\num{56}}
ROS systems usually provide a user interface (UI) for making the mission operator aware of the energy status of the robot.
Despite those are summary solutions, they are a key element for energy-efficient robots since they provide periodic updates about the energy conditions of the robot to roboticists and warn/alert them in case of critical energy issues.
In our study, we identify diverse types of UI: Organic Light-emitting Diode (OLED) and Liquid-crystal Display (LCD) displays (\datapoint{DP61} and \datapoint{DP509}), sound messages by using beeps and speakers (\datapoint{DP503} and \datapoint{DP515}), and dashboard widgets.
\todo[inline]{DataPoint for widgets.}
We also came across data points that discuss or implement specific components dedicated to operator notifications.
For instance, data point \datapoint{DP417} mentions a vocal package providing functionalities for alerting the operator via human-oriented audio alerts.

\theme{Reconfiguration of ROS components}{\num{23}}
ROS systems are distributed and rely on independent components that communicate by exchanging several messages at run-time.
Misconfigurations are common in ROS-based systems \cite{santos2017mining} and usually require fine adjustments and high expertise in the ROS communication middleware.
In most cases, the reconfiguration consists of modifying ROS launch files.
Another common solution is to release new ROS components after the robotic system is deployed, such as for data point \datapoint{DP258} that adds a battery guard, when the system is updated.
Moreover, there are cases where roboticists do not know how to launch a specific component, such as an energy monitoring package.
This is the case of data point \datapoint{DP183}, where a roboticist asks for help to publish some battery information as ROS messages: \q{I am trying to do is to create a package that executes some of these functions, for example, to read the voltage of the battery, and publish the data as ROS messages.}


\theme{Use of low-powered devices}{\num{22}}
In many cases, some components of a ROS-based system are deployed on an external Single-board Computer (SBC), as we can see in data points \datapoint{DP46}, \datapoint{DP48} and \datapoint{DP244}.
In the data point \datapoint{DP244}, a roboticist asks for help to launch ROS on a low-powered SBC.
He says to have chosen ASUS Xtion SBC due to \q{the lag of enough USB ports and the fact of less energy consumption.}
We also observe data points where the discussion heads to other low-powered devices.
This is the case of data point \datapoint{DP47}, where a roboticist discusses which laser scanner they should use for their robotic project.
One of the answers suggests a laser scanner with low power consumption: 
\q{many smaller robots use the Hokuyo URG-04LX, which is [...] cheaper, has lower power consumption but also much worse specs than the other two.}
The hardware needs to be purchased, and the price may vary a lot depending on their characteristics.
For instance, a LIDAR sensor for small projects costs less than \EUR{400}.
Therefore, knowing energy-efficient hardware in advance may also benefit economically.

\theme{Software improvements}{\num{35}}
This family of solutions represents improvements/fixes done exclusively on the software controlling the robot.
For example, data point \datapoint{DP84} is a code commit with a few line modifications correcting the robot behavior since a drone was not considering its battery level while taking off.
Another example is the data point \datapoint{DP219}, which corrects the battery status since it had not been updated in a graphical dashboard.
Note that both codes, before being corrected, may lead to serious energy issues due to wrong energy tracking.  
As an example of new software, we consider data point \datapoint{DP35}, which implements a new \emph{tf\_remapper} ROS node that according to developers helps to \q{stop wasting energy and CPU cycles}.
This node is used to deal with messages in a \emph{bag}, a file format to store logged data in ROS.
In this data point, the mention of energy waste is among many other technical details, which could have been overlooked without the extensive data analysis we conduct in this work.
Moreover, if roboticists do not adopt this module as a pattern, they keep relying on less energy-efficient ones, which as a consequence, might potentially lead to energy waste.

\theme{Other notable themes}{\num{71}}
Other notable solutions that our study reveals are: 
battery charge or recharge strategies (\num{14} occurrences);
the use of energy-efficient sensors (\num{11} occurrences);
simulation of real-world scenarios (\num{1}1 occurrences);
the software testing that either identifies energy-related bugs or validates corrections (\num{9} occurrences);
recommendation of alternative batteries (\num{8} occurrences);
documents that guide roboticists or developers (\num{7} occurrences);
deploy services remotely, such as in the Cloud (\num{6} occurrences); and
communication improvements, such as via serial ports or networked nodes (\num{5} occurrences).
Despite these solutions not being numerous, they are diverse and deserve roboticists' attention.
They comprise existing documentation that can provide information about how to implement new ROS nodes with energy efficiency in mind, together with dedicated software testing and simulation techniques. 
Moreover, remotely deploying services that assist or control robots is a good strategy since it keeps heavy computation away from the robot and, therefore, improves energy consumption \cite{pinto2017energy}.

\subsection{Mentioned quality attributes (\textbf{RQ4})}\label{sec:results_rq4}

In this RQ we investigated whether quality attributes are mentioned in the studied data points. 
From the \num{527} extracted data points, we noted that \num{134} of them (\num{25.4}\%) mention at least one quality attribute. The most common quality attributes are Usability, Performance, Reliability, Functional Stability, and Compatibility, although Usability is by far the most common one. Due to space constraints, in this section, we will briefly discuss the three most recurring quality attributes.

\theme{Usability}{\num{78}} Usability is a measure of how easy to use a user interface could be. Mentions to usability in our data points are mostly related to how to better show battery information to the robot operator. This could be achieved by, for instance, when one committer adds a script that highlights when the battery is at full capacity (\datapoint{DP20}).
Another way to enhance software usability is by providing accessibility. We noted many code changes aimed to use sound to indicate battery status. For instance, the following commit message suggested that ``\emph{Speak the remaining percentages of the battery if it's not charged}''; this particular pull-request (\datapoint{DP526}) was made at the project jsk\_robot with this goal. 


\theme{Performance}{\num{18}} Performance is about the process and techniques applied to software systems aimed to improve characteristics such as latency, throughput, or memory usage. In ROS Discourse, one roboticist reported that tf\_remap module is not efficient. As they reported: ``Have you ever run a tf\_remap on a complex system with tens of nodes subscribing TF? Have you looked at the CPU utilization? Stop wasting energy and CPU cycles by moving to tf\_remapper\_cpp'' (\datapoint{DP35}). 
Moreover, we found an issue on a ROS project that aims to convert files to videos to ease visualization. However, to improve the speed of the application (which was very slow in generating animations), ROS committers had to rely on multiprocessing and downsampling. These changes enable considerable speedup (\datapoint{DP73}).
Similarly, we found a commit that has the intention to disable a module instead of setting it to zero. According to the commit author, the intention was to ``avoid excessive power usage after shutdown.'' (\datapoint{DP447}).

\theme{Reliability}{\num{17}} Reliability is about performing a task consistently, to function without failure. 
To prevent that robot battery drains, which would cause the robot task to fail, we observed several discussions on how to make the robot to recharge its batteries autonomously (\ie docking (\eg \datapoint{DP206}).
Docking means that the robot stops whatever it is doing and drives back to a charging station.
Answers to this question were highly sophisticated since robots can do a myriad of tasks. 
For instance, we noticed a pull-request (\datapoint{DP252}) that implements a similar procedure. According to the author: ``\emph{When taking off, if the battery is low, should try to land; When reaching the goal, if the battery is low, should try to land}''. 



%% file: sections/discussion.tex
\section{Discussion}\label{sec:discussion}


In addition to the answers to RQ1-RQ4, in our analysis we also noticed recurrent reflection points about the state of the practice on energy-efficient robotics software.      

\noindent \textbf{Energy-related discussions are rare -- }
Only 0.002\% of the 339,563 considered artifacts contained energy-related discussions. This result confirms previous works that also found that these types of discussions are scarce among developers \cite{pang2015programmers}. 

\noindent \textbf{Managing energy-related issues is not trivial -- } 
In a non-negligible number of data points (\num{10} for the causes and \num{2} for the solutions), roboticists are seeking help about how to manage energy-related aspects of their software. As notable examples, in our dataset roboticists were seeking help for simulating battery usage in the Gazebo ROS simulator (\datapoint{DP42}), how to solve issues in the Kinect sensor due to it requiring extra power under some conditions \datapoint{DP115}, \etc 
This reinforces the importance of our study, especially our results for RQ3, which can support even experienced roboticists in solving energy-related issues in their ROS-based systems. 


\noindent \textbf{Energy consumption should be treated as a first-class quality attribute -- }
In some data points, we noticed that energy consumption is considered after an initial version of the system has been developed. This can be a problem in some cases since updating a ROS-based system after months of development might not be possible (\eg underwater robots); this phenomenon can be seen a form of energy-related technical debt \cite{verdecchia2020architectural}. As an example, in
\datapoint{DP44} a roboticist claims that \q{we would love to have low power consumption so that the robot can operate as long as possible [...] I would err towards the more expensive computer for an initial prototype so that you can get something running, and then work on optimizing the software to decrease the computing power required}.

\noindent \textbf{Energy efficiency as a bug -- }
In our analysis, we noticed that in \num{9} cases energy-efficiency mechanisms were actually the source of bugs and/or unexpected behavior within the system. Those bugs might be difficult to detect since they are linked to energy consumption in non-trivial ways. For example, in \datapoint{DP235} \q{when the turtle3 burger battery gets low [...] the robot stops moving but the local window [of the SLAM algorithm] keeps moving in the same direction at the same speed until the robot position on the local windows gets to the edge of the global map or over a known obstacle on the global map window}. Also, as mentioned in \datapoint{DP128}, some processes might go into sleep mode as an effect of a power save feature, potentially leading to unexpected availability/reliability issues.
We suggest roboticists (and researchers alike) carefully reflect on the possible consequences of having energy-saving modes in terms of system availability and reliability.

\noindent \textbf{ROS vs traditional energy efficiency -- } We noted that developing energy-efficient robots is not the same as developing a traditional energy-efficient app. This is partly due to the natural context of ROS-based software. For instance, robots are intrinsically battery-driven. When performing tasks such as driving or flying, ROS-based software should always monitor energy usage; if the battery is below a certain threshold, the robot should warn the user and/or possibly move back to the dock. This behavior brings additional energy challenges, since constantly monitoring the battery also incurs energy usage. Similarly, auto-docking is not always possible (given how far the robot is from the base). Therefore, managing a robot's energy consumption seems to be at least as challenging (but perhaps even more) when compared to traditional apps.

\noindent \textbf{Energy bugs in robotics software -- } Although in this work we do not perform a comprehensive analysis on energy bugs, we did discover some of them when investigating causes of energy-related issues. More interestingly, however, is the fact that these bugs are also domain-specific. We did not find traditional energy bugs such as the \texttt{loop bug} and the \texttt{no sleep bug}, which are among the most common ones in mobile apps~\cite{Pathak:EuroSys:2012,pathak2011bootstrapping}. For future work, we leave a fine-grained analysis of energy bugs in robotics software.

\noindent \textbf{UX skills might be on high demand for robotics software -- } We noticed in our data points that roboticists have to deal with a myriad of different types of UI. Many of our data points are related to improving the way that battery information is presented. Moreover, there also seems to be a lack of usability in ROS-based interfaces, evidenced by the high number of data points targeting this quality attribute. With that in mind, we believe that developers with UX experience would greatly benefit the ROS community.

\noindent \textbf{Robotics software tends to have unique energy-related issues -- }
Other studies in the literature investigated energy-related issues faced by software developers~\cite{pinto2014mining,moura2015mining}. Interestingly, there is very limited overlap between those and the ones faced by roboticists.  
For instance, in mobile apps development, recurrent causes of energy-related issues are (i) the misuse of wake locks  for keeping the CPU of the device active for long-running tasks~\cite{liu2016understanding} and (ii) the actual contents displayed on AMOLED displays~\cite{chen2013energy}, and (iii) networking overhead due to advertisement~\cite{pinto2014mining}. Those issues are not mentioned by roboticists when dealing with energy-related issues.



%% file: sections/threats.tex
\section{Threats to Validity}\label{sec:threats}


\noindent \textbf{External validity -- }
%
Despite the key role played by ROS in robotics software~\cite{quigley2009ros,sustainable_se_2020}, its coverage of a large variety of robots~\cite{ros2}, and its vibrant open-source community, we are aware that ROS-based projects 
may not cover all types of robots.
Our population data, however, is diverse ($\sim$\num{340}k data points, including \num{10} different robot families and \num{12} distinct robot capabilities). We extracted these data from multiple projects, covering a large number of contributors, commits, and types of robots.
Moreover, we selected the Git repositories for this study from the literature \cite{ICSE_SEIP_2020} and they already underwent a strict search and selection process, making us reasonably confident of their representativeness.


Our dataset has been built between February and April  2020. 
Some unsolved energy-related solutions may have been solved after that date, or new energy-related issues might have been reported. Since data from the analyzed repositories are changing constantly, it would be impractical to work on the most recent update all the time.
We mitigated this threat by including in our replication package a database with all data points we collected and investigated in this study.

This study reveals that only a small extent of our dataset is energy-related.
Nonetheless, we identify \num{10} energy-related causes and \num{14} energy-related solutions, which may impact decisions in robotics software engineering.
For instance, if software engineers are not aware of the impact of a monitoring rate, they will keep designing energy-inefficient robotics software.
Therefore, the scale of the study does not prevent us to identify critical energy-related practices for robotics software.

In this study, we do not consider mailing lists about ROS-based development. We deem this potential threat to validity as acceptable since the main communication channels for developers in the ROS ecosystem are ROS Answers and ROS Discourse~\cite{ROS_metrics}. 
Finally, our study comprises a relatively long period of time ($\sim$\num{11} years, \ie since ROS first releases);
to the best of our knowledge, this makes our study the one with the longest considered time span in the ROS domain.
Given that in this study we consider multiple data sources, there might be multiple data points about the same energy-related issue. 
For RQ1, we assume a balance on this software developer behaviour among years. 
RQ2 and RQ3 are qualitative, so repetitions can help in mitigating inconsistencies in the manual data analysis. 



\vspace{0.2cm}
\noindent \textbf{Internal validity -- }
%
%
%
We used a fixed set of keywords to search for energy-related data points while building the dataset (Section~\ref{sec:dataset}). 
Even if this search strategy requires relatively low effort, it proved to be highly effective in previous studies on energy-efficient software (\eg \cite{chowdhury2016characterizing,cruz2019catalog,malik2015going,pinto2014mining}). However, we know that it might lead to a high number of false positives and false negatives~\cite{bavota2016mining}. We mitigated this potential threat by (i) establishing the search keywords (and related taboo combinations) from the literature on energy-efficient software --- see Table~\ref{tab:paperswithterms} (\emph{Phase 1}), (ii) manually checking all 5,111 data points resulting from the search and removing all false positives (\emph{Phase 2}), and (iii) testing that the considered keywords are complete by manually checking a random sample of 400 data points without energy-related terms (\emph{Phase 3}). Also, three researchers were involved in the phases  mentioned above of our dataset construction process, following a known methodological procedure to avoid subjectivity~\cite{wohlin2012experimentation}.




Moreover, we answered RQ2 and RQ3 via thematic analysis. 
To mitigate possible biases due to subjectivity in the extraction of the codes and themes, we carefully followed the thematic analysis approach \cite{vaismoradi2013content}. 
Three researchers participated in the thematic analysis approach and all emerged themes were jointly revised until consensus was reached.
Finally, 
even though in some discussions the roboticists extensively elaborate on the impact at run-time in their projects, in this study we do not assess whether the emerged causes and solutions of energy-related issues actually impact the energy consumption of the robots. We deem this further investigation as out of scope for this specific study since (i) energy consumption is known to be heavily application-dependent \cite{pinto2017energy} and (ii) a definitive answer would require a dedicated experiment targeting each type of robot discussed in our dataset.

%% file: sections/related.tex
\section{Related Work}\label{sec:related}




\noindent \textbf{ROS-based software engineering research -- }The growing use of ROS in practice is also reflected by recent scientific publications targeting the ROS software ecosystem.
Fischer-Nielsen \etal~\cite{Fischer-Nielsen:SEIP:2020} studied dependency bugs (\ie bugs that appear when accessing a not available asset) on ROS repositories. Malavolta \etal~\cite{ICSE_SEIP_2020} studied 335 ROS repositories from an architectural standpoint and proposed a set of guidelines for the related software architecture design. Curan \etal~\cite{7140071} created a set of tools aimed at visualizing development metrics on ROS repositories to assess their maintenance health. None of these works, though, try to understand what are the energy-related problems that ROS developers have and how they cope with them, which is the goal of our research.


\noindent \textbf{Energy efficiency in the ROS ecosystem -- }Our work is not the first investigating energy efficiency in the ROS ecosystem. Swanborn and Malavolta~\cite{sustainable_se_2020} reviewed the existing body of research on energy efficiency in robotics software. They found 17 primary studies in the area. They observed that the first research work in this domain was published back in 1995~\cite{Barili:energy-saving-robots}, although the majority of the selected research works were published between 2012 and 2020. This shows the emerging character of the field. While most of the selected works are related to energy measurements and improvements, none of them addressed the analysis of developers' contributions on open source repositories. Thus, to the best of our knowledge, our work is the first to investigate how developers writing robot-oriented programs deal with energy issues.


\noindent \textbf{Mining energy-related data points -- } There is a significant number of research that exploits well-known developer repositories to gather energy-related information. Pinto \etal~\cite{pinto2014mining} studied the most popular questions (and related answers) about software energy consumption on StackOverflow.  Moura \etal~\cite{moura2015mining} performed similar work, but focused on energy-related code changes, \ie changes that developers do with the intention to reduce energy consumption. Inspired by the work of Pinto \etal~\cite{pinto2014mining} and Moura \etal~\cite{moura2015mining}, our work shares some of their research questions, but focuses on the ROS ecosystem. As such, we also expand these two contributions by exploring different Q\&A platforms and software repositories. 


Further, there are domain-specific research that share similar goals. For instance, some studies focus on mining energy-aware commits in the Android ecosystem~\cite{bao2016android}; mining energy-aware commits and pull-requests in Android and iOS ecosystems and then building a catalog of mobile energy patterns~\cite{cruz2019catalog}; or investigating if energy-aware commits have any impact on maintainability metrics~\cite{cruz2019maintainability}. These studies share a common limitation: since they rely on mining techniques, they do not measure the actual energy consumption data, and some of the solutions to energy-related problems might be limited to the understanding of the specific developers. 
Our work shares a similar limitation, too; however, we believe it has a lesser impact in our case, as the majority of our data points are based on textual discussions, and depend less on code changes.

%% file: sections/conclusions.tex
\section{Conclusions}\label{sec:conclusions}

Given the significant and increasing energy footprint of robots in various sectors, we analyzed the ROS ecosystem from various open-source channels; we quantified and characterized the main causes, solutions, and possible trade-offs of roboticists' energy-related issues.
%
%
Energy-related issues are scarce and the energy consumption is often addressed {\em after} delivery; at the same time, roboticists look for help to increase the energy efficiency of their robotics software. The practices resulting from our study offer the first step to help roboticists reuse solutions that already addressed energy efficiency, hence building increasingly-mature know-how about the development of energy-efficient robotics software. Our results also support researchers by providing the \textit{first comprehensive overview of the state of the practice on energy-related issues in robotics software}. Such an overview can help researchers in identifying impactful research directions for future contributions in software engineering and robotics.


%% file: main.bbl
\begin{thebibliography}{10}

\bibitem{robotics_roadmap}
{From Internet to robotics: A roadmap for US robotics: 2020 Edition}.
\newblock {\url{http://www.hichristensen.com/pdf/roadmap-2020.pdf}}, Oct 2020.
\newblock [Online; accessed 29. Oct. 2020].

\bibitem{robots_revenue}
{Industrial robotics market revenue worldwide 2025 {$\vert$} Statista}.
\newblock
  \url{https://www.statista.com/statistics/760207/worldwide-industrial-robotics-market-revenue},
  Oct 2020.
\newblock [Online; accessed 29. Oct. 2020].

\bibitem{BibEntry2020May}
{Release Notes for Onboard SDK 3.9 - DJI Onboard SDK Documentation}.
\newblock
  \url{https://developer.dji.com/onboard-sdk/documentation/appendix/releaseNotes.html},
  May 2020.
\newblock [Online; accessed 6. Jan. 2021].

\bibitem{ROS_metrics}
{ROS Community Metrics}.
\newblock \url{http://wiki.ros.org/Metrics}, Jul 2020.
\newblock [Online; accessed 28. Oct. 2020].

\bibitem{ISO25010}
I.~25010:2011.
\newblock {Systems and software engineering - Systems and software Quality
  Requirements and Evaluation (SQuaRE) - System and software quality models}.
\newblock \url{https://www.iso.org/standard/35733.html}, 2011.

\bibitem{replication_package}
{Anonymous}.
\newblock {MSR 2021 Replication Package}.
\newblock
  \url{https://github.com/S2-group/msr-2021-robotics-green-practices-replication-package},
  2021.

\bibitem{bao2016android}
L.~Bao, D.~Lo, X.~Xia, X.~Wang, and C.~Tian.
\newblock How android app developers manage power consumption?-an empirical
  study by mining power management commits.
\newblock In {\em 2016 IEEE/ACM 13th Working Conference on Mining Software
  Repositories (MSR)}, pages 37--48. IEEE, 2016.

\bibitem{Barili:energy-saving-robots}
A.~{Barili}, M.~{Ceresa}, and C.~{Parisi}.
\newblock Energy-saving motion control for an autonomous mobile robot.
\newblock In {\em 1995 Proceedings of the IEEE International Symposium on
  Industrial Electronics}, volume~2, pages 674--676 vol.2, 1995.

\bibitem{bavota2016mining}
G.~Bavota.
\newblock Mining unstructured data in software repositories: Current and future
  trends.
\newblock In {\em 2016 IEEE 23rd International Conference on Software Analysis,
  Evolution, and Reengineering (SANER)}, volume~5, pages 1--12. IEEE, 2016.

\bibitem{calero2013systematic}
C.~Calero, M.~F. Bertoa, and M.~{\'A}. Moraga.
\newblock A systematic literature review for software sustainability measures.
\newblock In {\em 2013 2nd international workshop on green and sustainable
  software (GREENS)}, pages 46--53. IEEE, 2013.

\bibitem{chang2012energy}
Q.~Chang, G.~Xiao, S.~Biller, and L.~Li.
\newblock Energy saving opportunity analysis of automotive serial production
  systems (march 2012).
\newblock {\em IEEE Transactions on Automation Science and Engineering},
  10(2):334--342, 2012.

\bibitem{chen2013energy}
X.~Chen, Y.~Chen, Z.~Ma, and F.~C. Fernandes.
\newblock How is energy consumed in smartphone display applications?
\newblock In {\em Proceedings of the 14th Workshop on Mobile Computing Systems
  and Applications}, pages 1--6, 2013.

\bibitem{chowdhury2016characterizing}
S.~A. Chowdhury and A.~Hindle.
\newblock Characterizing energy-aware software projects: Are they different?
\newblock In {\em Proceedings of the 13th International Conference on Mining
  Software Repositories}, pages 508--511, 2016.

\bibitem{ICSE_2017_TB}
F.~Ciccozzi, D.~D. Ruscio, I.~Malavolta, P.~Pelliccione, and J.~Tumova.
\newblock Engineering the software of robotic systems.
\newblock In {\em Proceedings of the 39th International Conference on Software
  Engineering Companion}, pages 507--508. IEEE Press, May 2017.

\bibitem{CK}
J.~Cohen.
\newblock Weighted kappa: Nominal scale agreement provision for scaled
  disagreement or partial credit.
\newblock {\em Psychological bulletin}, 70(4):213, 1968.

\bibitem{cruz2019catalog}
L.~Cruz and R.~Abreu.
\newblock Catalog of energy patterns for mobile applications.
\newblock {\em Empirical Software Engineering}, 24(4):2209--2235, 2019.

\bibitem{cruz2019maintainability}
L.~Cruz, R.~Abreu, J.~Grundy, L.~Li, and X.~Xia.
\newblock Do energy-oriented changes hinder maintainability?
\newblock In {\em 2019 IEEE International Conference on Software Maintenance
  and Evolution (ICSME)}, pages 29--40. IEEE, 2019.

\bibitem{thematic_analysis}
D.~S. {Cruzes} and T.~{Dyba}.
\newblock Recommended steps for thematic synthesis in software engineering.
\newblock In {\em 2011 International Symposium on Empirical Software
  Engineering and Measurement}, pages 275--284, Sep. 2011.

\bibitem{7140071}
W.~{Curran}, T.~{Thornton}, B.~{Arvey}, and W.~D. {Smart}.
\newblock Evaluating impact in the ros ecosystem.
\newblock In {\em 2015 IEEE International Conference on Robotics and Automation
  (ICRA)}, pages 6213--6219, May 2015.

\bibitem{jss_2019}
P.~Estefo, J.~Simmonds, R.~Robbes, and J.~Fabry.
\newblock The robot operating system: Package reuse and community dynamics.
\newblock {\em Journal of Systems and Software}, 151:226--242, 2019.

\bibitem{fereday2006demonstrating}
J.~Fereday and E.~Muir-Cochrane.
\newblock Demonstrating rigor using thematic analysis: A hybrid approach of
  inductive and deductive coding and theme development.
\newblock {\em International journal of qualitative methods}, 5(1):80--92,
  2006.

\bibitem{Fischer-Nielsen:SEIP:2020}
A.~{Fischer-Nielsen}, Z.~{Fu}, T.~{Su}, and A.~{Wąsowski}.
\newblock The forgotten case of the dependency bugs : On the example of the
  robot operating system.
\newblock In {\em 2020 IEEE/ACM 42nd International Conference on Software
  Engineering: Software Engineering in Practice (ICSE-SEIP)}, pages 21--30,
  2020.

\bibitem{ros2}
B.~Gerkey.
\newblock Why {ROS} 2?
\newblock \url{https://design.ros2.org/articles/why_ros2.html}, 2019.

\bibitem{hao2013estimating}
S.~Hao, D.~Li, W.~G. Halfond, and R.~Govindan.
\newblock Estimating mobile application energy consumption using program
  analysis.
\newblock In {\em 2013 35th international conference on software engineering
  (ICSE)}, pages 92--101. IEEE, 2013.

\bibitem{hernandez2015}
S.~H. Juan and F.~H. Cotarelo.
\newblock Multi-master ros systems.
\newblock Technical report, 2015.

\bibitem{kate2018phys}
S.~Kate, J.-P. Ore, X.~Zhang, S.~Elbaum, and Z.~Xu.
\newblock Phys: probabilistic physical unit assignment and inconsistency
  detection.
\newblock In {\em Proceedings of the 2018 26th ACM Joint Meeting on European
  Software Engineering Conference and Symposium on the Foundations of Software
  Engineering}, pages 563--573, 2018.

\bibitem{li2020detecting}
X.~Li, Y.~Yang, Y.~Liu, J.~P. Gallagher, and K.~Wu.
\newblock Detecting and diagnosing energy issues for mobile applications.
\newblock In {\em Proceedings of the 29th ACM SIGSOFT International Symposium
  on Software Testing and Analysis}, pages 115--127, 2020.

\bibitem{lidwell2010universal}
W.~Lidwell, K.~Holden, and J.~Butler.
\newblock {\em Universal principles of design}.
\newblock Rockport Pub, 2010.

\bibitem{liu2016understanding}
Y.~Liu, C.~Xu, S.-C. Cheung, and V.~Terragni.
\newblock Understanding and detecting wake lock misuses for android
  applications.
\newblock In {\em Proceedings of the 2016 24th ACM SIGSOFT International
  Symposium on Foundations of Software Engineering}, pages 396--409, 2016.

\bibitem{MSR_2021_architectural_tactics}
I.~Malavolta, K.~Chinnappan, S.~Swanborn, G.~Lewis, and P.~Lago.
\newblock { Mining the ROS ecosystem for Green Architectural Tactics in
  Robotics and an Empirical Evaluation }.
\newblock In {\em Proceedings of the 18th International Conference on Mining
  Software Repositories, {MSR}}, page To appear, New York, NY, May 2021. ACM.

\bibitem{ICSE_SEIP_2020}
I.~Malavolta, G.~Lewis, B.~Schmerl, P.~Lago, and D.~Garlan.
\newblock How do you architect your robots? state of the practice and
  guidelines for {ROS}-based systems.
\newblock In {\em {ACM/IEEE} International Conference on Software Engineering},
  2020.

\bibitem{malik2015going}
H.~Malik, P.~Zhao, and M.~Godfrey.
\newblock Going green: An exploratory analysis of energy-related questions.
\newblock In {\em IEEE/ACM Working Conference on Mining Software Repositories},
  pages 418--421, 2015.

\bibitem{matalonga2019greenhub}
H.~Matalonga, B.~Cabral, F.~Castor, M.~Couto, R.~Pereira, S.~M. de~Sousa, and
  J.~P. Fernandes.
\newblock Greenhub farmer: real-world data for android energy mining.
\newblock In {\em 2019 IEEE/ACM 16th International Conference on Mining
  Software Repositories (MSR)}, pages 171--175. IEEE, 2019.

\bibitem{meike2013energy}
D.~Meike, M.~Pellicciari, and G.~Berselli.
\newblock Energy efficient use of multirobot production lines in the automotive
  industry: Detailed system modeling and optimization.
\newblock {\em IEEE Transactions on Automation Science and Engineering},
  11(3):798--809, 2013.

\bibitem{moghaddam2018self}
F.~A. Moghaddam, P.~Lago, and I.~C. Ban.
\newblock Self-adaptation approaches for energy efficiency: a systematic
  literature review.
\newblock In {\em 2018 IEEE/ACM 6th International Workshop on Green And
  Sustainable Software (GREENS)}, pages 35--42. IEEE, 2018.

\bibitem{moura2015mining}
I.~Moura, G.~Pinto, F.~Ebert, and F.~Castor.
\newblock Mining energy-aware commits.
\newblock In {\em 2015 IEEE/ACM 12th Working Conference on Mining Software
  Repositories}, pages 56--67. IEEE, 2015.

\bibitem{pang2015programmers}
C.~Pang, A.~Hindle, B.~Adams, and A.~E. Hassan.
\newblock What do programmers know about software energy consumption?
\newblock {\em IEEE Software}, 33(3):83--89, 2015.

\bibitem{pathak2011bootstrapping}
A.~Pathak, Y.~C. Hu, and M.~Zhang.
\newblock Bootstrapping energy debugging on smartphones: a first look at energy
  bugs in mobile devices.
\newblock In {\em Proceedings of the 10th ACM Workshop on Hot Topics in
  Networks}, pages 1--6, 2011.

\bibitem{Pathak:EuroSys:2012}
A.~Pathak, Y.~C. Hu, and M.~Zhang.
\newblock Where is the energy spent inside my app? fine grained energy
  accounting on smartphones with eprof.
\newblock In {\em Proceedings of the 7th ACM European Conference on Computer
  Systems}, EuroSys '12, page 29–42, New York, NY, USA, 2012. Association for
  Computing Machinery.

\bibitem{NL_robots}
R.~Paulissen, S.~Kalisingh, J.~Scholtes, A.~van Geldrop, and A.-L. Hoftijzer.
\newblock {Robotics in the Netherlands}.
\newblock {\em Netherlands Foreign Investment Agency (NFIA) report}, 2016.

\bibitem{pinto2017energy}
G.~Pinto and F.~Castor.
\newblock Energy efficiency: a new concern for application software developers.
\newblock {\em Communications of the ACM}, 60(12):68--75, 2017.

\bibitem{pinto2014mining}
G.~Pinto, F.~Castor, and Y.~D. Liu.
\newblock Mining questions about software energy consumption.
\newblock In {\em Proceedings of the 11th Working Conference on Mining Software
  Repositories}, pages 22--31, 2014.

\bibitem{procaccianti2013energy}
G.~Procaccianti, S.~Bevini, and P.~Lago.
\newblock Energy efficiency in cloud software architectures.
\newblock In {\em EnviroInfo}, pages 291--299, 2013.

\bibitem{procaccianti2015systematic}
G.~Procaccianti, P.~Lago, and S.~Bevini.
\newblock A systematic literature review on energy efficiency in cloud software
  architectures.
\newblock {\em Sustainable Computing: Informatics and Systems}, 7:2--10, 2015.

\bibitem{ros}
M.~Quigley, K.~Conley, B.~Gerkey, J.~Faust, T.~Foote, J.~Leibs, R.~Wheeler, and
  A.~Y. Ng.
\newblock {ROS}: an open-source robot operating system.
\newblock In {\em ICRA workshop on open source software}, volume~3, page~5.
  Kobe, Japan, 2009.

\bibitem{quigley2009ros}
M.~Quigley, K.~Conley, B.~Gerkey, J.~Faust, T.~Foote, J.~Leibs, R.~Wheeler, and
  A.~Y. Ng.
\newblock Ros: an open-source robot operating system.
\newblock In {\em ICRA workshop on open source software}, volume~3, page~5.
  Kobe, Japan, 2009.

\bibitem{quigley2015programming}
M.~Quigley, B.~Gerkey, and W.~D. Smart.
\newblock {\em Programming Robots with ROS: a practical introduction to the
  Robot Operating System}.
\newblock O'Reilly Media, Inc., 2015.

\bibitem{santos2017mining}
A.~Santos, A.~Cunha, N.~Macedo, R.~Arrais, and F.~N. Dos~Santos.
\newblock Mining the usage patterns of ros primitives.
\newblock In {\em 2017 IEEE/RSJ International Conference on Intelligent Robots
  and Systems (IROS)}, pages 3855--3860. IEEE, 2017.

\bibitem{shull2007guide}
F.~Shull, J.~Singer, and D.~I. Sj{\o}berg.
\newblock {\em Guide to advanced empirical software engineering}.
\newblock Springer, 2007.

\bibitem{sustainable_se_2020}
S.~Swanborn and I.~Malavolta.
\newblock Energy efficiency in robotics software: A systematic literature
  review.
\newblock In {\em 35th IEEE/ACM International Conference on Automated Software
  Engineering Workshops (ASEW '20)}, pages 137--144. ACM, 2020.

\bibitem{vaismoradi2013content}
M.~Vaismoradi, H.~Turunen, and T.~Bondas.
\newblock Content analysis and thematic analysis: Implications for conducting a
  qualitative descriptive study.
\newblock {\em Nursing \& health sciences}, 15(3):398--405, 2013.

\bibitem{verdecchia2020architectural}
R.~Verdecchia, P.~Kruchten, P.~Lago, and I.~Malavolta.
\newblock Building and evaluating a theory of architectural technical debt
  insoftware-intensive systems.
\newblock {\em Journal of Systems and Software}, page 110925, 2021.

\bibitem{wohlin2012experimentation}
C.~Wohlin, P.~Runeson, M.~H{\"o}st, M.~Ohlsson, B.~Regnell, and A.~Wessl{\'e}n.
\newblock {\em Experimentation in Software Engineering}.
\newblock Computer Science. Springer, 2012.

\end{thebibliography}
